\documentclass[10pt,conference]{IEEEtran}
\IEEEoverridecommandlockouts
\usepackage{cite}
\usepackage{amsmath,amssymb,amsfonts}
\usepackage{graphicx}
\usepackage{textcomp}
\usepackage[dvipsnames]{xcolor}
\usepackage{xspace}
\usepackage{cases}
\usepackage{hyperref}
\usepackage{pifont}
\usepackage{subcaption}
\usepackage{balance}
\usepackage{framed}
\usepackage{formatting/shortcuts}
\usepackage{mathtools}
\usepackage[noend]{algpseudocode}
\usepackage{algorithm}

\usepackage{tikz}

\makeatletter
\newcommand{\linebreakand}{%
  \end{@IEEEauthorhalign}
  \hfill\mbox{}\par
  \mbox{}\hfill\begin{@IEEEauthorhalign}
}
\makeatother
\def\BibTeX{{\rm B\kern-.05em{\sc i\kern-.025em b}\kern-.08em
    T\kern-.1667em\lower.7ex\hbox{E}\kern-.125emX}}
\begin{document}

\title{Scalable Quantitative Verification For Deep Neural Networks}
\author{\IEEEauthorblockN{Teodora Baluta}
\IEEEauthorblockA{\textit{National University of Singapore} \\
teobaluta@comp.nus.edu.sg}
\and
\IEEEauthorblockN{Zheng Leong Chua}
\IEEEauthorblockA{\textit{Independent Researcher} \\
czl@iiyume.org}
\and
\IEEEauthorblockN{Kuldeep S. Meel}
\IEEEauthorblockA{\textit{National University of Singapore} \\
meel@comp.nus.edu.sg}
\linebreakand
\IEEEauthorblockN{Prateek Saxena}
\IEEEauthorblockA{\textit{National University of Singapore} \\
prateeks@comp.nus.edu.sg}
}

\maketitle

\begin{abstract}
  Despite the functional success of deep neural networks (DNNs), their
  trustworthiness remains a crucial open challenge. To address this
  challenge, both testing and verification techniques have been
  proposed. But these existing techniques provide either scalability
  to large networks or formal guarantees, not both. In this paper, we
  propose a scalable {\em quantitative verification} framework for
  deep neural networks, i.e., a test-driven approach that comes with
  formal guarantees that a desired probabilistic property is
  satisfied. Our technique performs enough tests until soundness of a
  formal probabilistic property can be proven. It can be used to
  certify properties of both deterministic and randomized DNNs.  We
  implement our approach in a tool called \codename\footnote{The name is a pun
  on {\em proverò} (I will prove it) and {\em pro-vero} (pro-truth) in Italian.
Code and benchmarks are available at \url{teobaluta.github.io/provero}} and apply it in
  the context of certifying adversarial robustness of DNNs. In this
  context, we first show a new {\em attack-agnostic} measure of
  robustness which offers an alternative to purely attack-based
  methodology of evaluating robustness being reported today. Second,
  \codename provides certificates of robustness for large DNNs, where
  existing state-of-the-art verification tools fail to produce
  conclusive results. Our work paves the way forward for verifying
  properties of distributions captured by real-world deep neural
  networks, with provable guarantees, even where testers only have
  black-box access to the neural network.
\end{abstract}

\section{Introduction}
The past few years have witnessed an increasing adoption of deep neural networks
(DNNs) in domains such as autonomous
vehicles~\cite{bojarski2016end,spielberg2019neural,litman2017autonomous},
drones~\cite{julian2016policy} or
robotics~\cite{giusti2016machine,yun2017action}, where mispredictions can have
serious long-term consequences. Robustness, privacy, and fairness
have emerged as central concerns to be addressed for safe adoption of
DNNs~\cite{carlini2017towards,shokri2017membership,lecuyer2018certified,papernot2018sok,song2019privacy,amodei2016concrete,mehrabi2019survey}.
Consequently, there has been a growing attention to testing and verification of neural
networks for properties of interest.

To establish that the resulting DNNs have the desired properties, a
large body of prior work has focused on techniques based on empirical
testing~\cite{tian2018deeptest,sun2018concolic,xie2019deephunter,gopinath2018deepsafe} or
specialized attack
vectors~\cite{madry2017towards,uesato2018adversarial,papernot2016transferability,athalye2018obfuscated,shokri2017membership,gopinath2019symbolic}.
While such techniques are useful, they {\em do not} rigorously
quantify how sure we can be that the desired property is true after
testing.

In contrast to testing approaches, formal verification seeks to
provide rigorous guarantees of correctness.  Inspired by the success
of model checking in the context of hardware and software
verification, the earliest formal verification methodologies in the
context of deep neural networks focused on qualitative verification,
i.e., whether a system satisfies a given specification.  Prior work in
this category has been following the model checking paradigm wherein a
given DNN is encoded as a model using constraints grounded in a chosen
theory. Then a satisfiability solver (often modulo the chosen theory)
is invoked to check if there exists an execution of the system that
violates the given
specification~\cite{pulina2010abstraction,ehlers2017formal,narodytska2017verifying,katz2017reluplex}.

The proposed techniques in this category appear to have three limitations.
Firstly, they require white-box access to the models and specialized procedures
to transform the DNNs to a specification, limiting their generality. Secondly,
the performance of the underlying feasibility solver degrades severely with the
usage of non-linear constraints, leading to analyses that do not scale to larger models.
Thirdly, prior techniques are limited to deterministic neural networks, while
extensive research effort has been invested in designing randomized DNNs,
especially to enhance
robustness~\cite{dhillon2018stochastic,lecuyer2018certified,cohen2019certified}.

Such qualitative verification considers only two scenarios: either a
DNN satisfies the property, or it does not. However, neural networks
are stochastically trained, and more importantly, they may run on
inputs drawn from an unknown distribution at inference
time. Properties of interest are thus often probabilistic and defined
over an input distribution (e.g., fairness~\cite{mehrabi2019survey} or
robustness to distributional
changes~\cite{amodei2016concrete}). Hence, qualitative verification is
unsuitable for such properties.

An alternative approach is to check {\em how often} a property is satisfied by a
given DNN under a given input distribution.  More specifically, one can assert
that a DNN satisfies a property $\psi$ with a desirably high probability
$1-\delta$. Unlike ad-hoc testing, {\em quantitative
verification}~\cite{BSSMS19} aims to provide {\em soundness}, i.e., when it
confirms that $\psi$ is true with probability $p$, then the claim can be
rigorously deduced from the laws of probability. 
For many practical
applications, knowing that the chance of failure is controllably small suffices
for deployment.  For instance, it has been suggested that road safety standards
for self-driving cars specify sufficiently low failure rates of the perceptual
sub-systems, against which implementations can be
verified~\cite{koopman2019safety,kalra2016driving,thorn2018framework}.
Further, we show the role of quantitative verification in the specific context
of adversarial robustness (see Section~\ref{app:robustness}).

In this paper, we present a new quantitative verification algorithm for DNNs
called \codename, tackling the following problem: Given a logical property
$\psi$ specified over a space of inputs and outputs of a DNN and a numerical
threshold $\theta$, decide whether $\psi$ is true for less than $\theta$
fraction of the inputs.
\codename is a procedure that achieves the above goal with proven soundness:
When it halts with a \yes or \no decision, it is correct with probability $1 -
\delta$ and within approximation error $\eta$ to the given $\theta$. The
verifier can control the desired parameters $(\eta, \delta)$, making them
arbitrarily close to zero. That is, the verifier can have controllably high
certainty about the verifier's output, and $\theta$ can be arbitrarily precise
(or close to the ground truth).  The lower the choice of $(\eta, \delta)$ used
by the verifier, the higher is the running time.

\codename is based on sampling, and it makes only one assumption---the ability
to take independent samples from the space over which $\psi$ is
defined\footnote{For non-deterministic DNNs, the procedure assumes that the
  randomization used for the DNN is independent of its specific input.}.  This
  makes the verification procedure considerably general and stand-alone. The
  verifier only needs black-box access to the DNN, freeing it up from assuming
  anything about the internal implementation of the DNNs. The DNN can be
  deterministic or from a general family of non-deterministic DNNs. This allows
  checking probabilistic properties of deterministic DNNs and of randomization
  procedures defined over DNNs.

  Our paper makes the following contributions:
\begin{itemize}
\item We present a new quantitative verification algorithm for neural
  networks. The framework is fairly general: It assumes only black-box
  access to the model being verified, and assumes only the ability to
  sample from the input distribution over which the property is
  asserted.

\item We implement our approach in a tool called \codename that embodies sound
  algorithms and scales quantitative verification for adversarial robustness to
  large real-world DNNs, both deterministic and randomized, within $1$ hour.

\item In the context of certifying adversarial robustness, our
  empirical evaluation presents a new attack-agnostic measure of
  robustness and shows that \tool can produce certificates with high
  confidence on instances where existing state-of-the-art qualitative
  verification does not provide conclusive results.

\end{itemize}

\section{Application: Adversarial Robustness}
\label{sec:application}

For concreteness, we apply our approach to verifying the robustness of neural
networks. In proving robustness, the analyst has to provide a space of inputs
over which the robustness holds. Often, this space is defined by all inputs that
are within a {\em perturbation bound} $\epsilon$ of a given input $\vec{x}$ in
the $L_p$ norm~\cite{goodfellow2014explaining}.  Different distance norms have
been used such as $L_0$, $L_1$, $L_2$ and $L_\infty$. The $L_p$ norm is defined
as $\norm{\vec{x}-\vec{x}'}_p=(|x'_1 - x_1|^p + |x'_2 - x_2|^p + \ldots + |x'_n
- x_n|^p)^{1/p}$.  A neural network $f$ is defined to be robust with respect to
a given input $\vec{x}$ if
$  \forall \vec{x}'$ such that $\norm{\vec{x}-\vec{x}'}_p < \epsilon,$ we have $ f(\vec{x}) =
  f(\vec{x}')$.

For a given neural network $f$ and input point $\vec{x}$, there always exists
some perturbation size beyond which there are one or more adversarial samples.
We refer to this minimum perturbation with non-zero adversarial examples as
$\epsilon_{min}$, which is the ground truth the security analyst wants to
know. Attack procedures are best-effort methods which find upper bounds for
$\epsilon_{min}$ but cannot provably show that these bounds are
tight~\cite{tramer2017ensemble,carlini2016hidden,athalye2018obfuscated}.
Verification procedures aim to prove the absence of adversarial examples below
a given bound, i.e., they can establish lower bounds for $\epsilon_{min}$.  We
call such verified lower bounds $\epsilon_{verf}$. Most verifiers proposed to
date for robustness checking are qualitative, i.e., given a perturbation
size $\epsilon_{verf}$, they output whether adversarial examples are absent within
$\epsilon_{verf}$. If the verification procedure is sound and outputs \yes, then it is
guaranteed that there are no adversarial examples within $\epsilon_{verf}$,
i.e., the robustness property is satisfied.  When the verifier says \no, if the
verifier is complete, then it is guaranteed that there are indeed adversarial
examples within $\epsilon_{verf}$.  If the verifier is incomplete and
prints \no, the result is inconclusive.

Let us introduce a simple quantitative measure of robustness called the {\em
adversarial density}. Adversarial density is the fraction of inputs around a
given input $\vec{x}$ which are adversarial examples. We explain why adversarial
density is a practically useful quantity and much easier to compute for large
DNNs than $\epsilon_{min}$. We can compute perturbation bounds below which the
adversarial density is non-zero but negligibly small, and we empirically show
these bounds are highly correlated with estimates of $\epsilon_{min}$ obtained
by state-of-the-art attack methods.

\begin{figure}
  \centering
  \begin{subfigure}{0.45\columnwidth}
  \includegraphics[width=\linewidth]{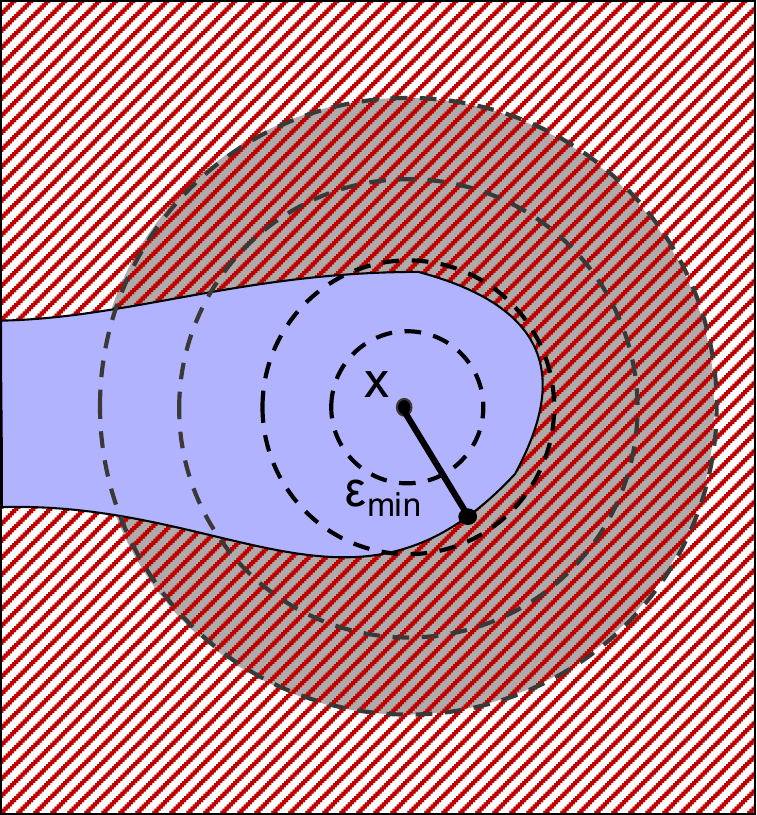}
  \end{subfigure}
  \centering
  \begin{subfigure}{0.45\columnwidth}
  \includegraphics[width=\linewidth]{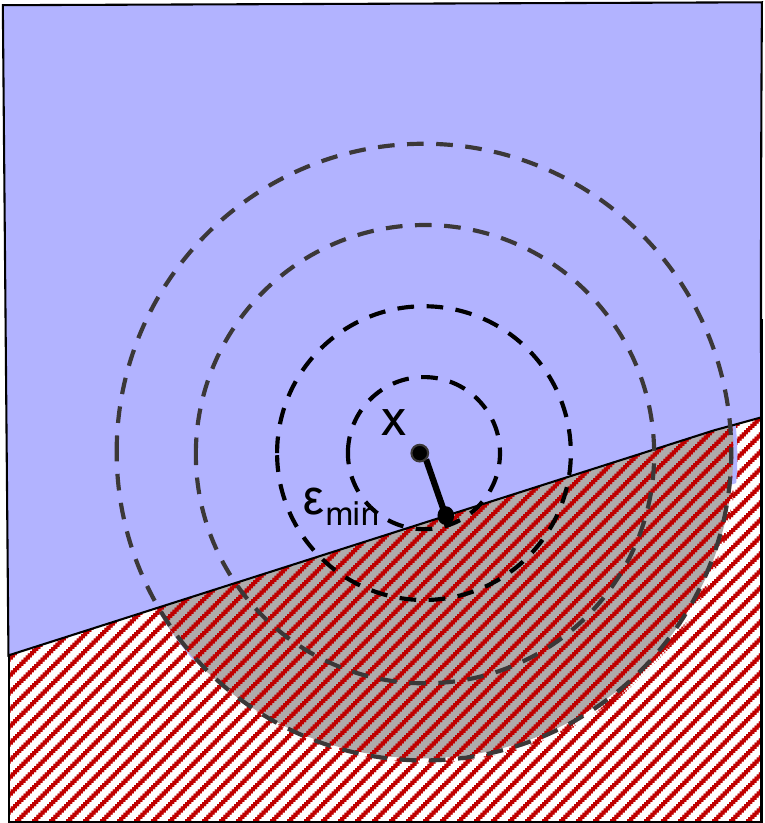}
  \end{subfigure}
  \caption{The decision boundaries of two binary classifiers $f_1$
    (left) and $f_2$ (right) around  an input $\vec{x}$ are shown. 
    The correct classification region for $\vec{x}$ is shown in purple (solid),
    while the incorrect classification region is shown in red (hashed). The concentric circles show the
    equidistant points from $\vec{x}$ in $L_2$-norm drawn up to a bound
    $\epsilon$.  The classifier
    $f_1$ has a better (larger) minimum perturbation  than $f_2$, but has
    a worse (larger) adversarial density because a majority of points within
    $\epsilon$ distance of $\vec{x}$ are in the incorrect classification region.
    Therefore, the smoothened version of $f_2$ will classify
    $\vec{x}$ correctly, while the smoothened $f_1$ will not. Picking the base
    classifier with the better adversarial density, rather than minimum
    perturbation, will lead to better accuracy in a smoothening defense.
  }
  \label{fig:adv_density}
\end{figure}

\subsection{Minimum Perturbation vs. Density}
\label{app:robustness}

It is reasonable to ask why adversarial density is relevant at all for security
analysis. After all, the adversary would
exploit the weakest link, so the minimum perturbation size $\epsilon_{min}$ is
perhaps the only quantity of interest. 
We present concrete instances where adversarial density is relevant.

First, we point to randomized smoothing as a defense technique, which
has provable guarantees of adversarial
robustness~\cite{lecuyer2018certified,dhillon2018stochastic,cohen2019certified,wong2018provable}.
The defense uses a ``smoothed'' classifier $g$ that averages the
behavior of a given neural net $f$ (called the base classifier) around
a given input $\vec{x}$. More specifically, given a base classifier
$f$, the procedure samples perturbations of $\vec{x}$ within
$\epsilon$ from a specific noise distribution and outputs the most
likely class $c$ such that $argmax_{c \in \mathcal{Y}}
Pr[f(\vec{x}+\epsilon) = c]$. Notice that this procedure computes the
probability of $f$ returning class $c$---typically by counting how
often $f$ predicts class $c$ over many samples---rather than
considering the "worst-case" example around $\vec{x}$.  Said another
way, these approaches estimate the adversarial density for each output
class under some input distribution. Therefore, when
selecting between two base classifiers during training, we should pick
the one with the smallest adversarial density for the correct 
class, irrespective of their minimum adversarial perturbation size.

To illustrate this point, in Figure~\ref{fig:adv_density} we show two 
DNNs $f_1$ and $f_2$, as potential candidates for the base classifier in a
randomized smoothing procedure.  Notice that $f_1$ has a  better (larger)
$\epsilon_{min}$ than $f_2$. However, more of the inputs within the
$\epsilon$-ball of the input $\vec{x}$ (inside the red hashed circle) are
classified as the wrong label by $f_1$ in comparison to $f_2$. Thus, a
smoothed classifier with $f_1$ as a base classifier would misclassify $\vec{x}$
where the smoothed classifier with base $f_2$ would classify correctly.  
This explains why we should choose the
classifier with the smaller adversarial density rather than one based
on the minimum perturbation because the smoothing process is not susceptible to
worst-case examples by its very construction.
This motivates why computing adversarial density is useful for adversarial
robustness defenses.

Second, we point out that estimating minimal perturbation bounds has been a
difficult problem. Attack procedures, which provide an upper bound for
$\epsilon_{min}$, are constantly evolving.  This makes robustness evaluations
attack-specific and a moving target. On the other hand, qualitative verification
techniques can certify that the DNN has no adversarial examples below a certain
perturbation, which is a lower bound on the adversarial
perturbation~\cite{katz2017reluplex,tjeng2017evaluating,singh2018fast,dvijotham2018dual}.
However, these analyses do not scale well with deep networks and can lead either
to timeouts or inconclusive results for large real-world DNNs.  Furthermore,
they are {\em white-box}, requiring access to the model internals and work only
for deterministic neural networks.
We show in this work that verifying adversarial density bounds is easy to
compute even for large DNNs. We describe procedures that require only black-box
access, the ability to sample from desired distributions and
hence are attack-agnostic.

In particular, we show an empirical attack-agnostic metric for estimating robustness of a given
DNN and input $\vec{x}$ called {\em adversarial hardness}. It is highest
perturbation bound for which the adversarial density is below a suitably low
$\theta$. We can search empirically for the highest perturbation bound
{\advhard}, called the adversarial hardness, for which a sound quantitative
certifier says \yes when queried with
$(f,\vec{x},${\advhard}$,\theta,\delta,\error)$---implying that $f$ has suitably
low density of adversarial examples for perturbation bounds
below {\advhard}.

Adversarial hardness is a measure
of the difficulty of finding an adversarial example by uniform
sampling.
Surprisingly, we find that this measure strongly
correlates with perturbation bounds produced by prominent  white-box
attacks (see Section~\ref{sec:eval}). 
Given this strong correlation, we can effectively use adversarial hardness as a
proxy for perturbation sizes obtained from specific attacks,
when comparing the relative robustness of two DNNs. 

We caution readers that adversarial hardness is a quantitative
measure and technically
different from $\epsilon_{min}$, the distance to the nearest adversarial
example around $\vec{x}$. But both these measures provide complementary
information about the concentration of adversarial examples near a
perturbation size.

\section{Problem Definition}
\label{sec:prob}

We are given a neural network and a space of its inputs over which we
want to assert a desirable property $\prop$ of the outputs of the
network. Our framework allows one to check whether $\prop$ is true for
some specified ratio $\thresh$ of all possible values in the specified
space of inputs. For instance, one can check whether most inputs, a
sufficiently small number of inputs, or any other specified constant
ratio of the inputs satisfies $\prop$. The specified ratio parameter
$\thresh$ is called a {\em threshold}.

Formally, let $\prop(\samplespace, f, \Phi)\in\{0,1\}$ be a property function over a neural
network $f$, a subset of all possible inputs to
the neural network $\samplespace$ and user-specified parameters $\Phi$. 
We assume that we can efficiently draw samples from any probability distribution
over $\samplespace$ that the analyst desires.
For a given distribution $D$ over $\samplespace$, let $p_{D} =\mathbf{E}_{x\sim
D}[\prop(x,f,\Phi)]$. $p_D$ can be viewed as the probability that $\prop$
evaluates to $1$ for $x$ sampled from $\samplespace$ according to $D$. When
clear from context, we omit $D$ and simply use $p$ to refer to $p_{D}$.

Ideally, one would like to design an algorithm that returns `Yes' if $p \leq
\thresh$ and 'No' otherwise. Such exact quantification is intractable, so we are
instead interested in an algorithm $\mathcal{A}$ that returns `Yes' if $p \leq
\thresh$ and 'No' otherwise, with two controllable approximation parameters
$(\error,\delta)$. The procedure should be theoretically {\em sound}, ensuring
that when it produces `Yes' or `No' results, it is correct with probability at
least $1-\delta$ within an additive bound on its error $\error$ from the
specified threshold $\thresh$.  Specifically, we say that algorithm
$\mathcal{A}$ is {\em sound} if:

\begin{align*}
&Pr[\mathcal{A}(\prop, \thresh, \error, \delta)~ \text{returns \todo{\yes}}~|~p \leq \thresh] \geq 1 - \delta
\\
&Pr[\mathcal{A}(\prop, \thresh, \error, \delta)~ \text{returns \todo{\no}}~|~p > \thresh +\error] \geq 1 - \delta
\end{align*}

The analyst has arbitrary control over the confidence $\delta$ about
$\mathcal{A}$'s output correctness and the precision $\error$ around the
threshold. These values can be made arbitrarily small approaching
zero, increasing the runtime of $\mathcal{A}$.  The soundness guarantee is
useful---it rigorously estimates how many inputs in $\samplespace$ satisfy
$\prop$, serving as a quantitative metric of satisfiability.

The presented framework makes very few assumptions. It can work with
any specification of $\samplespace$, as long as there is an interface to be able
to sample from it (as per any desired distribution) efficiently.
The neural network $f$ can be any deterministic function. In fact, it can be any
``stateless'' randomized procedure, i.e., the function evaluated on a particular
input does not use outputs produced from prior inputs. This general class of
neural networks includes, for instance, Bayesian neural
networks~\cite{houthooft2016curiosity} and many other ensembles of neural network
architectures~\cite{barber1998ensemble}.
The framework permits specifying all properties of fairness, privacy, and
robustness highlighted in recent prior
work~\cite{BSSMS19,narodytska2019assessing}, for a much broader class of DNNs.

Our goal is to present sound and scalable algorithms for quantitative
verification, specifically targeting empirical performance for
quantitatively certifying robustness of DNNs. The framework assumes
black-box access to $f$, which can be deterministic or
non-deterministic. Our techniques can directly check qualitative
certificates produced from randomized robustness-enhancing defenses,
one example of which is the recent work called
PixelDP~\cite{lecuyer2018certified} (see Section~\ref{sec:pixeldp-eval}).

\section{Approach Overview}

\subsection{Sampling}

Given a property $\prop$ over a sampleable input space $\samplespace$
and a neural network $f$, our approach works by sampling $\nbsamples$ times
independently from $\samplespace$. We test $f$ on each sample as
input.
Let $\rndX_i$ be a 0-1 random variable denoting the result of the
trial with sample $i$, where $\rndX_i = 1$ if the $\prop(x,f,\Phi)$ is true
and $\rndX_i = 0$ otherwise.  Let $\rndX$ be the random variable
denoting the number of trials in $\rndX_1, \rndX_2, \dots, \rndX_{\nbsamples}$
for which the property is true.
Then, the standard Chernoff bounds (see~\cite{mitzenmacher2017probability})
given below form the main workhorse underlying our algorithms:

\begin{lemma}
\label{lemma:chernoff}
Given independent 0-1 random variables $X_1,\hdots,X_{\nbsamples}$, let $X =
\sum_{i=1}^{\nbsamples}{X_i}$, $\mu=\frac{\mathbf{E}[\rndX]}{\nbsamples}$, and
$\hat{p}=\frac{\rndX}{N}$.  For $0 < \eta < 1$:
\begin{align*}
Pr[\hat{p} \geq \mu + \error] \leq e^{\frac{-N\error^2}{3\mu}}\\ 
Pr[\hat{p} \leq \mu - \error] \leq e^{\frac{-N\error^2}{2\mu}}
\end{align*}
\end{lemma}

Note that the probability $p$ we are interested in bounding in our quantitative
verification framework is exactly $\mu$ in Lemma~\ref{lemma:chernoff}.  More
specifically, the probability depends on the neural network and distribution
over the inputs, $p=\mathbf{E}_{x\sim D}[\prop(x,f,\Phi)]$, where $D$
is a distribution over $\samplespace$.
Using a framework based on sampling and Chernoff bounds admits
considerable generality.  The only assumption in applying the
Lemma~\ref{lemma:chernoff} is that all samples are independent. If the
neural network does not compute information during one trial (or
execution under one sample) and use it in another trial, as is the case for all
neural networks we study, trials will be independent.
For any deterministic neural network, all samples are drawn
independently and identically distributed in $\samplespace$, so
Chernoff bounds are applicable.
For randomized DNNs, we can think of the $i^{th}$ trial as evaluating
a potentially different neural network (sampled from some distribution
of functions) on the given sample. Here, the output random variables
may not be identically distributed due to the randomization used by
the neural network itself. However, Lemma~\ref{lemma:chernoff} can
still be used even for non-identically distributed trials but independent.

We discuss an estimation-based strategy that applies Chernoff bounds in a
straight-forward manner next. Such a solution has high sample complexity for
quantitative verification of adversarial robustness. We then propose
hypothesis-based solutions which are {\em sound} and have much better empirical
sample complexity for real-world DNNs. Our proposed algorithms still rely only
on Chernoff-style bounds, but are carefully designed to internally vary
parameters (on which Chernoff bounds are invoked) to reduce the number of
samples needed to dispatch the asserted property.

\subsection{An Estimation-based Solution}
\label{sec:baseline}

One way to quantitatively verify a property through sampling is to
directly apply Chernoff bounds to the empirical estimate of the mean
$\hat{p}$ in $\nbsamples$ trials. The solution is to take $\nbsamples >
\frac{12 \cdot ln\frac{1}{\delta}}{\error^2}$ samples, and decide \yes
if $\hat{p} \leq \thresh + \frac{\error}{2}$ and \no if $\hat{p} >
\thresh + \frac{\error}{2}$.
This is a common estimation approach, for instance used in the prediction step
in the certified defense mechanism PixelDP~\cite{lecuyer2018certified}.
By Lemma~\ref{lemma:chernoff}, one can show that $\hat{p}$ is within
$\pm\frac{\eta}{2}$ additive error of $p$ with confidence higher than
$1-\delta$. Therefore, the procedure is sound since $Pr[p \not\in [\hat{p} -
\frac{\error}{2}, \hat{p} + \frac{\error}{2})] \leq \delta$ by
Lemma~\ref{lemma:chernoff}, for all $0 \leq p \leq 1$.

In this solution, the number of samples increase quadratically with
decreasing $\error$. For example, if the
$\thresh=0.1,\error=10^{-3},\delta=0.01$, directly applying Chernoff
bounds will require over $55 \times 10^{6}$ samples.  Even for an
optimized architecture such as
BranchyNet~\cite{teerapittayanon2016branchynet} that reports $70.9$ {\em ms} on
average inference time per sample for a ResNet ($152$ layers) the estimation
approach would take more than $1083$ GPU compute hours. This is a prohibitive
cost.  For randomized DNNs, which internally compute expectations, the runtime
of the estimation baseline approach can be even larger. For example, the
randomization used in PixelDP can have $3-42\times$ inference overhead compared
to deterministic DNNs~\cite{lecuyer2018certified}.

Our work provides new algorithms that utilize much fewer samples on average.  In
the example of BranchyNet mentioned above, if the true probability $p$ is $0.3$, our
approach requires $4246$ samples to return a `No' answer with confidence
greater than $0.99$.
The main issue with the estimation algorithm is that it does not utilize
the knowledge of the given $\thresh$ in deciding the number of samples
it needs.

\subsection{Our Approach}

The number of samples needed for Chernoff bounds depend on how far is
the true probability $p$ that we are trying to bound from the given
threshold $\theta$. Intuitively, if $p$ and $\theta$ are far apart in
the interval $\lbrack 0,1 \rbrack$, then a small number of samples are
sufficient to conclude with high certainty that $p \leq \theta$ or $p >
\theta + \eta$ (for small $\eta$). The estimation approach takes the
same number of samples irrespective of how far $p$ and $\theta$
are. Our algorithms terminate quickly by checking for
``quick-to-test'' hypothesis early, yielding the sample complexity
comparable to the estimation only in the worst case.

We propose new algorithms, the key idea of which is to use cheaper (in
sample complexity) hypothesis tests to decide \yes or \no early.
Given the threshold $\thresh$ and the error $\error$, the high-level
idea is to propose alternative hypotheses on the left side of
$\thresh$ and on the right side of $\thresh+\error$. We choose the
hypotheses and a sampling procedure such that if any of the hypotheses
on the left side of $\thresh$ are true, then we can return
\yes. Similarly, if any of the hypotheses on the right side of
$\thresh+\error$ are true, then we can return \no. Thus, we can
potentially return much faster when \todo{the true probability} $p$ is further
from \todo{the threshold} $\thresh$.

\begin{algorithm}[t]
  \caption{\metaprovero($\thresh,\error, \delta$)}
  \label{alg:meta}
  \begin{algorithmic}[1]
    \While{\colorbox{BurntOrange}{\texttt{cond}}}
    \State \colorbox{BurntOrange}{\texttt{pick}} $\thresh_1<\thresh_2 \leq
    \thresh$ \label{alg:choose-left-interval}
    \State ans = \tester($\thresh_1, \thresh_2, \delta$)
    \label{alg:tester-call1}
    \If{ans == \yes} \Return \yes \EndIf
    \State \colorbox{BurntOrange}{\texttt{pick}} $\thresh_2>\thresh_1 \geq
  \thresh + \error$ \label{alg:choose-right-interval}
    \State ans = \tester($\thresh_1, \thresh_2, \delta$)
    \label{alg:tester-call2}
    \If{ans == \no} \Return \no \EndIf
  \EndWhile
  \State \Return \tester($\thresh, \thresh + \error, \delta$)
\end{algorithmic}
\end{algorithm}

\begin{algorithm}[t]
  \caption{\tester($\thresh_1,\thresh_2,\delta$)}
  \label{alg:tester}
  \begin{algorithmic}[1]
  \State $\nbsamples=\frac{(\sqrt{3\theta_1} + \sqrt{2\theta_2})^2}{(\theta_2 -
\theta_1)^2}ln\frac{1}{\delta}$
  \State $\error_1 = (\theta_2 -
  \theta_1)\Big(1+\sqrt{\frac{2}{3}\frac{\theta_2}{\theta_1}}\Big)^{-1}$
  \State $\error_2 = \theta_2 - \theta_1 - \error_1$
  \State \texttt{sample} $\nbsamples$ times
  \State $\hat{p} = \frac{1}{\nbsamples}\sum_{i=1}^\nbsamples{X_i}$ \Comment $X_i$ - samples
  that satisfy the property
  \If{$\hat{p} \leq \thresh_1 + \error_1$} \Return \yes \EndIf
  \If{$\hat{p} > \thresh_2 - \error_2$} \Return \no \EndIf
  \State \Return None
  \end{algorithmic}
\end{algorithm}

The overall meta-level structure of our algorithms is simple and follows
Algorithm~\ref{alg:meta}, called \metaprovero. \todo{In
lines~\ref{alg:choose-left-interval} and~\ref{alg:choose-right-interval},
we pick the alternative hypotheses on the left and right of the
given threshold $\thresh$ respectively.} We sample a certain number of
samples, estimate the ratio $\hat{p}$ (by invoking \tester in
\todo{lines~\ref{alg:tester-call1} and~\ref{alg:tester-call2}}),
and check if we can prove that conditions involving \todo{the alternative
intermediate} thresholds ($\thresh_1$ and $\thresh_2$) are satisfied with the desired input
parameters $(\eta,\delta)$ using Chernoff bounds. If the check succeeds, the
algorithm can return \yes or \no; otherwise, the process repeats until a
condition which guarantees soundness is met.

The internal thresholds are picked so as to soundly {\em prove} or {\em refute}
that $p$ lies in certain ranges in $\lbrack 0,1 \rbrack$.
\todo{
This is done by testing certain {\em intervals} $p$ is (or is not) in. For
instance, \metaprovero tests a pair of hypotheses $p \leq \thresh_1$ and $p >
\thresh_2$ simultaneously (line \ref{alg:tester-call1}).
}
Notice that for the intervals on the left side of $\thresh$ ($\thresh_1 <
\thresh_2 \leq \thresh$, \todo{chosen in line \ref{alg:choose-left-interval} in
Alg.~\ref{alg:meta})}, the call to \tester can result in proving
that $p\leq\thresh_1$ with desired confidence $\delta$ and error tolerance
$\eta$.  In this case, since $\thresh_1 < \thresh$, we will have proven the
original hypothesis $p\leq\thresh$ and the algorithm can return \yes soundly.
We call such intervals, which are to the left of $\thresh$, as {\em
  proving intervals}.
Conversely, {\em refuting intervals} are on the right side of $\thresh+\error$.
\todo{The choice of $\thresh_1$ and $\thresh_2$ on
line~\ref{alg:choose-right-interval} in Alg.~\ref{alg:meta} is such that they
are larger than $\thresh + \error$}. When we can prove that $p > \thresh_2$,
\todo{i.e., the \tester call on line~\ref{alg:tester-call2} returns \no, then
we can soundly return \no because $\thresh_2 > \thresh+\error$ implies $p
> \thresh+\error$.}

\usetikzlibrary{patterns}
\usetikzlibrary{arrows}

\makeatletter
\tikzset{%
        hatch distance/.store in=\hatchdistance,
        hatch distance=5pt,
        hatch thickness/.store in=\hatchthickness,
        hatch thickness=5pt
        }
\pgfdeclarepatternformonly[\hatchdistance,\hatchthickness]{north east hatch}%
    {\pgfqpoint{-1pt}{-1pt}}%
    {\pgfqpoint{\hatchdistance}{\hatchdistance}}%
    {\pgfpoint{\hatchdistance-1pt}{\hatchdistance-1pt}}%
    {
        \pgfsetcolor{\tikz@pattern@color}
        \pgfsetlinewidth{\hatchthickness}
        \pgfpathmoveto{\pgfqpoint{0pt}{0pt}}
        \pgfpathlineto{\pgfqpoint{\hatchdistance}{\hatchdistance}}
        \pgfusepath{stroke}
    }
\makeatother

\begin{figure}
  \centering
\resizebox{0.8\columnwidth}{!}{
\begin{tikzpicture}[ultra thick]
\fill[pattern=north east hatch,pattern color=green,hatch distance=8mm, hatch
thickness=.5pt] (-4,0) rectangle node[midway,above] {} (-2.4,2.4);
\fill[pattern=north east hatch,pattern color=pink, hatch distance=3mm, hatch
thickness=2pt] (-2.2,0) rectangle node[midway,above] {} (4,2.4);
\fill[color=red] (-2.4,0) rectangle (-2.2,2.4);

\draw (-4,-0.2) -- node[pos=-0.5] {0} (-4,0.2);
\draw (4,-0.2) -- node[pos=-0.5] {1} (4,0.2);
\begin{scope}[thick]
\draw (-2.4,0) -- node[above=1.4cm,left=0.1cm] {$p\leq\theta$} (-2.4,2.4);
\draw (-2.2,0) -- node[above=1.4cm,right=0.01cm] {$p>\theta+\error$} (-2.2,2.4);
\draw (0.5,0)[color=blue] -- node[pos=-0.5,color=blue] {$p$} (0.5, 1);
\draw (-3.5,0) -- node[pos=-0.5] {$\theta_{1_l}$} (-3.5, 0.5);
\draw (-3,0) -- node[pos=-0.5] {$\theta_{2_l}$} (-3, 0.5);
\draw (-1.3,0) -- node[pos=-0.5] {$\theta_{1_r}$} (-1.3, 0.5);
\draw (0,0) -- node[pos=-0.5] {$\theta_{2_r}$} (0, 0.5);
\end{scope}
\draw (-4,0) -- (4,0);

\end{tikzpicture}
}
\caption{Example run of \metaprovero($0.1,10^{-3},0.01$) given that $p=0.3$. On
  the left-hand side, \metaprovero cannot prove that $p \leq \thresh$ since $p$
  is greater than left-side $\thresh_{2_l}$.  Next, \metaprovero chooses the refuting
  interval on the right-hand side. Since $p > \thresh_{2_r}$, \tester returns
  \no and \metaprovero can prove that $p > \thresh + \error$.}
\label{fig:example-run}
\end{figure}

\todo{In Fig.~\ref{fig:example-run}, we consider an example run of
  \metaprovero given that the true probability $p=0.3$ (highlighted in blue) and
  the input parameters are $\thresh=0.1,\error=10^{-3},\delta=0.01$. \metaprovero picks the
  proving interval $(\thresh_{1l}, \thresh_{2l})=(0.03, 0.06)$ on the left of $\thresh$ and calls
  \tester which returns \no. This means that the true probability is greater than
  $\thresh_{1l}$ with high confidence. \metaprovero then picks an
  interval $(\thresh_{1r}, \thresh_{2r})=(0.15, 0.25)$ on the right-side of
$\thresh+\error$. Here \tester returns with confidence higher than $1-\delta$ that
$p > \thresh_{2r}$. Since $\thresh_{2r} > \thresh + \error$ we can conclude that
the true probability is greater than $\thresh$ with error $\error$.}

The key building block of this algorithm is the \tester sub-procedure
(Algorithm~\ref{alg:tester}), which employs sampling to check hypotheses.
Informally, the \tester does the following: Given two intermediate
thresholds, $\thresh_1$ and $\thresh_2$, if the true probability $p$
is either smaller than $\thresh_1$ or greater than $\thresh_2$, it
returns \yes or \no respectively with high confidence.  If $p \in
(\thresh_1, \thresh_2)$ then the tester does not have any guarantees.
Notice that a single invocation of the \tester checks {\em two hypotheses
simultaneously}, using one set of samples. The number of samples needed are
proven sufficient in Section~\ref{sec:tester}.

One can directly invoke \tester with $\thresh_1=\thresh$ and
$\thresh_2=\thresh + \error$ but that might lead to a very large
number of samples, $\mathcal{O}(1/\error^2)$. Thus, the key challenge
is to judiciously call the \tester on hypotheses with smaller sample
complexity such that we can refute or prove faster in most cases.  To
this end, notice that \metaprovero leaves out two algorithmic design
choices: stopping conditions and the strategy for choosing the proving
and refuting hypotheses (highlighted in Alg.~\ref{alg:meta}).
We propose and analyze an adaptive {\em binary-search}-style algorithm where we
change our hypotheses based on the outcomes of our sampling tests
(Section~\ref{sec:adapt}).
We show that our proposed algorithm using this strategy is sound.
When $p$ is extremely close to $\theta$, these algorithms
are no worse than estimating the probability, requiring roughly the same number
of samples.

\section{Algorithms}

We provide an adaptive algorithm for quantitative certification that narrows the
size of the intervals in the proof search, similar to a binary-search strategy
(Section~\ref{sec:adapt}). This algorithm build on the base of the \tester primitive
which we explain in Section~\ref{sec:tester}.

\subsection{The \bincert Algorithm}
\label{sec:adapt}

We propose an algorithm \bincert (Algorithm~\ref{alg:adapt}) where instead of
fixing the intervals beforehand we narrow our search by halving the intervals.
The user-specified input parameters for \bincert are the threshold
$\thresh$, the error bound $\error$ and the confidence parameter $\delta$. The
interval creation strategy is off-loaded to the \createint procedure outlined in
Algorithm~\ref{alg:createint}.  The interval size $\alpha$ is initially set to
the largest possible as \tester would require less samples on wider intervals
(Algorithm~\ref{alg:createint},
lines~\ref{line:first-create-int-start}-\ref{line:first-create-int-end}). Then, the \bincert algorithm calls
internally the procedure \createint to create {\em proving} intervals (on the
left side of $\thresh$, Alg.~\ref{alg:adapt}, line~\ref{line:adaptive-left-int}) and {\em refuting}
intervals (on the right side of $\thresh + \error$, Alg.~\ref{alg:adapt},
line~\ref{line:adaptive-right-int}).
Note that for the refuting intervals, we keep the left-side fixed, $\thresh_1 =
\thresh + \error$ and for the proving intervals we keep the right-side fixed,
i.e., $\thresh_2 = \thresh$ . For each iteration of \bincert, the
strategy we use is to halve the intervals by moving the outermost thresholds
closer to $\thresh$ (Alg.~\ref{alg:createint},
lines~\ref{line:move-threshold-start}-\ref{line:move-threshold-end}).
For these intermediate hypotheses, \tester checks if it can prove or
disprove the assert $p \leq \thresh$ (lines~\ref{line:left-tester-call}
and~\ref{line:right-tester-call}).
It continues to do so alternating the proving and refuting hypotheses until the
size of both intervals becomes smaller than the error bound $\error$
(line~\ref{line:bincert-final}).
If only on one side of the threshold \createint returns intervals with size
$\alpha > \error$, \bincert checks those hypotheses.
If the size of the proving and refuting intervals returned by \createint is
$\error$, then the final check is directly invoked on $(\theta, \theta+\error)$
and returned to the user (line~\ref{line:bincert-final-tester}).

The algorithm \bincert returns \yes or \no with soundness guarantees as defined
in Section~\ref{sec:prob}. We give our main theorem here and defer the proof to
Section~\ref{sec:analysis}:
\begin{theorem}\label{thm:adapt-guarantee}
  For an unknown value $p \in [0,1]$, a given threshold $\theta\in[0,1],
  \error\in(0,1), \delta\in(0,1]$, \bincert  returns a \yes or \no with the
  following guarantees:
\begin{align*}
  &Pr[\text{\bincert returns \todo{\yes}} ~|~ p \leq \theta] \geq 1 - \delta \\
  &Pr[\text{\bincert returns \todo{\no}} ~|~ p > \theta + \error] \geq 1 - \delta
\end{align*}
\end{theorem}

  \begin{algorithm}[H]
\caption{\bincert$(\thresh, \error, \delta)$}
\label{alg:adapt}
\begin{algorithmic}[1]
  \State $\thresh_{1_l}=\thresh_{{2}_l}=0$
  \State $\thresh_{1_r}=\thresh_{{2}_r}=0$
  \State $n = 3 + max(0,log(\frac{\thresh}{\error})) + max(0,log(\frac{1-\thresh-\error}{\error}))$
  \State $\delta_{min}=\delta / n$ \Comment minimum confidence per call to \tester
  \While{True}
    \State $\thresh_{1_l}, \thresh_{{2}_l}$ = \Call{CreateInterval}{$\thresh,
    \thresh_{1_l}, \thresh_{{2}_l},\error$, True}\label{line:adaptive-left-int}
    \If {$\thresh_{{2}_l} - \thresh_{1_l} > \error$}
      \State ans = \Call{Tester}{$\thresh_{1_l}, \thresh_{2_l}$,
      $\delta_{min}$}\label{line:left-tester-call}
      \If {ans == \todo{\yes}}
      \Return \todo{\yes}
      \EndIf
    \EndIf
    \State $\thresh_{1_r}, \thresh_{{2}_r}$ = \Call{CreateInterval}{$\thresh,
    \thresh_{1_r}, \thresh_{{2}_r}, \error,$ False}\label{line:adaptive-right-int}
    \If {$\thresh_{{2}_r} - \thresh_{1_r} > \error$}
    \State ans = \Call{Tester}{$\thresh_{1_r}, \thresh_{2_r}$,
    $\delta_{min}$}\label{line:right-tester-call}
      \If{ans == \todo{\no}}
      \Return \todo{\no}
      \EndIf
    \EndIf
    \If {$\thresh_{{2}_r} - \thresh_{1_r} \leq \error$ and $\thresh_{{2}_l} - \thresh_{1_l} \leq \error$} \label{line:bincert-final}
      \State \Return \Call{Tester}{$\thresh, \thresh+\error$, $\delta_{min}$} \label{line:bincert-final-tester}
    \EndIf
  \EndWhile
\end{algorithmic}
\end{algorithm}
\hfill

\vspace{-2.5em}
  \begin{algorithm}[H]
\caption{\createint($\thresh, \thresh_1, \thresh_2, \error$, left)}
\label{alg:createint}
\begin{algorithmic}[1]
  \If {$\thresh == 0$ and left}
  \Return $(\thresh, \thresh + \error)$
  \EndIf\label{line:first-create-int-start}
  \If {$\thresh_1 == 0$ and $\thresh_2 == 0$ and left}:
    \Return ($0, \thresh$)
  \EndIf
  \If {$\thresh_1 == 0$ and $\thresh_2 ==0$ and $\neg$left}:
    \Return ($\thresh+\error, 1$)
  \EndIf\label{line:first-create-int-end}

  \State $\alpha = \thresh_2 - \thresh_1$
  \If {left}\label{line:move-threshold-start}
    \State\Return ($\thresh_2$ - max($\error, \alpha /2$), $\thresh_2$)
  \EndIf
  \State\Return ($\thresh_1, \thresh_1$ + max($\error,
  \alpha/2$))\label{line:move-threshold-end}
\end{algorithmic}
\end{algorithm}

\subsection{\textsc{Tester} Primitive}
\label{sec:tester}

The tester takes as input two thresholds $\theta_1, \theta_2$ such that
$\theta_1 < \theta_2$ and confidence parameter $\delta$ and returns
`Yes' when $p \leq \theta_1$ with confidence higher than $1 - \delta$ and `No'
when $p > \theta_2$ with confidence higher than $1 - \delta$. If $p \in
(\theta_1, \theta_2]$ the \tester returns without guarantees.

The procedure for implementing the \tester is simple. Following the procedure
outlined in Algorithm~\ref{alg:tester}, we take
$\nbsamples=\frac{(\sqrt{3\theta_1}+\sqrt{2\theta_2})^2}{(\theta_2 -
\theta_1)^2}ln\frac{1}{\delta}$ number of independent samples and estimate the
ratio of these 0-1 trials as $\hat{p}$. The \tester returns \yes if $\hat{p}
\leq \thresh_1 + \error_1$ and \no if $\hat{p} > \thresh_2 - \error_2$ where
$\error_1$ and $\error_2$ are error parameters (lines 2 and 3). If $\thresh_1 +
\error_1 < \hat{p} < \thresh_2 - \error_2$, our implementation returns 'None'.
The following lemma establishes the soundness of the tester, and follows
directly from applying Chernoff bounds on $\hat{p}$.

\begin{lemma}
  \label{lemma:tester} Given the thresholds ($\thresh_1,\thresh_2$) and
  confidence parameter $\delta$, \tester has following soundness
  guarantees:
\begin{align*}
  &Pr[\text{\tester returns \yes} ~|~ p \leq \thresh_1] \geq 1 - \delta\\
  &Pr[\text{\tester returns \no} ~|~ p > \thresh_2] \geq 1 - \delta
\end{align*}
\end{lemma}

\begin{proof}
  The proof follows directly the Chernoff bounds.
\end{proof}

\pgfmathdeclarefunction{gauss}{3}{%
  \pgfmathparse{1/(#3*sqrt(2*pi))*exp(-((#1-#2)^2)/(2*#3^2))}%
}

\begin{figure}
  \centering
\resizebox{0.8\columnwidth}{!}{
\begin{tikzpicture}
\begin{axis}[
    no markers
  , domain=-7.5:25.5
  , samples=100
  , ymin=0
  , axis lines*=left
  , xlabel=
   , every axis y label/.style={at=(current axis.above origin),anchor=south}
  , every axis x label/.style={at=(current axis.right of origin),anchor=west}
  , height=3cm
  , width=10cm
  , xtick=\empty
  , ytick=\empty
  , enlargelimits=false
  , clip=false
  , axis on top
  , grid = major
  , hide y axis
  , hide x axis
  ]

\addplot[name path=A, ultra thick,restrict x to domain=-5:6] {gauss(x, 0, 1.75)};
\pgfmathsetmacro\valueA{gauss(0, 0, 1.75)}
\draw [dashed, thick] (axis cs:0, 0) -- (axis cs:0, \valueA+0.1);
\node[below] at (axis cs:0, -0.02)  {\Large {$\thresh_{1}$}};
\draw[thick, ] (axis cs:-0.0, -0.01) -- (axis cs:0.0, 0.01);

\addplot[name path=B, ultra thick,restrict x to domain=0.5:10] {gauss(x, 5, 1.3)};
\draw [dashed, thick] (axis cs:6, 0) -- (axis cs:6, \valueA+0.1);
\node[below] at (axis cs:6, -0.02)  {\Large {$\thresh_{2}$}};
\draw[thick ] (axis cs:6, -0.01) -- (axis cs:6, 0.01);

\draw[|-|,thick, black] (axis cs:-5, 0) -- (axis cs:10, 0);
\node[below] at (axis cs:-5, -0.02)  {0};
\node[below] at (axis cs:10, -0.02)  {1};

\node[below] (t) at (axis cs: 2.8, -0.03) {\Large$t$};
\node[above, black] at (axis cs: 2, 0.3) {$\error_1$};
\draw[thick, black,<->] (axis cs:2.7, 0.3) -- (axis cs:6, 0.3);
\draw[thick, black,<->] (axis cs:0, 0.3) -- (axis cs:2.7, 0.3);
\node[above, black] at (axis cs: 5, 0.3) {$\error_2$};
\draw [dashed, thick] (axis cs:2.7, 0) -- (axis cs:2.7, \valueA+0.1);
\node[below] (bounded) at (axis cs: 3.3, -0.2)  {each bounded by $\delta$};
\node[above] (yes) at (axis cs: -2.0, -0.3)  {return \yes};
\node[above] (no) at (axis cs: 8, -0.3)  {return \no};
\draw[->, black, thick] (yes) edge (axis cs: -1.0, 0.1);
\draw[->, black, thick] (no) -- (axis cs: 5, 0.1);
\draw[->, black, thick] (axis cs: 2.2, -0.2) -- (axis cs: 2.3, 0.03);
\draw[->, black, thick] (axis cs: 3.2, -0.2) -- (axis cs: 3.2, 0.03);

\path[name path=xaxis]
  (0,0) -- (11,0); %
\path[ name path=lower,
intersection segments={
  of=A and B,
  sequence=R1 -- L2,
}
];

\addplot [fill=red, fill opacity=0.75] fill between [of=xaxis and lower];

\addplot[pattern=north west lines] fill between[of=A and lower];
\addplot[pattern=north east lines] fill between[of=B and lower];

\end{axis}

\end{tikzpicture}
}
\captionof{figure}{\tester considers the boundary
$t=\thresh_1+\error_1=\thresh_2-\error_2$ that allows to distinguish
$p\leq\thresh_1$ or $p > \thresh_2$.}
\label{fig:tester}
\end{figure}

Using the estimated probability $\hat{p}$, \tester returns \yes if $\hat{p} \leq
\thresh_1+\error_1$ and \no if $\hat{p} \geq \thresh_2-\error_2$ with
probability greater than $1-\delta$. Otherwise, it returns `None'.
We want to choose values $\error_1,\error_2
\in (0,1)$ such that $Pr[\hat{p} \geq \theta_1 +  \error_1 ~|~ p \leq \thresh_1]
\leq \delta$ and $Pr[\hat{p} \leq \theta_2 - \error_2 ~|~ p > \thresh_2] \leq
\delta$.

To derive the minimum number of samples for the estimated $\hat{p}$, the key
idea is to use one set of samples to check two hypotheses, $p \leq \thresh_1$
and $p > \thresh_2$, simultaneously. To do so, we find a point $t \in
(\thresh_1, \thresh_2)$ which serves as a decision boundary for the
estimate probability $\hat{p}$. 
We illustrate this in Figure~\ref{fig:tester}: it shows $(t,\eta_1,\eta_2)$ and
the two probability distributions for $\hat{p}$ given $p=\theta_1$ and
$p=\theta_2$, respectively. The distributions for the case $p < \theta_1$ will
be shifted further to the left, and the case $p > \theta_2$ will be shifted
further to the right; so these are extremal distributions to consider.
It can be seen that $t$ is chosen such that $Pr[\hat{p} > t | p \leq \theta_1]$
as well as $Pr[\hat{p} < t | p > \theta_2]$ are bounded (shaded red) by
probability $\delta$.
Using the additive Chernoff bounds (Lemma~\ref{lemma:chernoff}), for a
given $\thresh_1$ and $\thresh_2$, the number of samples
$\nbsamples$ is the maximum of $3\theta_1
\frac{1}{\error_1^2}ln\frac{1}{\delta}$ and $2\theta_2
\frac{1}{\error_2^2}ln\frac{1}{\delta}$.

Taking the maximum ensures that the probabilities of the both the
hypotheses, $p<\theta_1$ and $p>\theta_2$, are being simultaneously
upper bounded. 
We leave the detail analysis for the supplementary material.

\section{Soundness}
\label{sec:analysis}

In this section, we prove that our proposed algorithm satisfies soundness as
defined in Section~\ref{sec:prob}. 
\bincert uses the \tester primitive on certain intervals in sequence. Depending
on the strategy, the size of the intervals and the order of testing them
differs.  But, the algorithm terminates immediately if the \tester returns \yes
on a proving interval or \no on a refuting one.  The meta-algorithm captures
this structure on line 2-5 and \bincert algorithm instantiates this general
structure.  When none of these optimistic calls to \tester are successful, the
algorithm makes a call to the \tester on the remaining interval in the worst
case. Given the same basic structure of algorithms as per the meta-algorithm
\metaprovero, we now prove the following key theorem:

\begin{lemma}
\label{lemma:soundness}
Let $E$ be the event that the algorithm $\cal{A} \in \{\bincert\}$ fails, then $Pr[E] \leq \delta$.
\end{lemma}

\begin{proof} 
Fix any input to
$\cal{A}$, and consider the execution of $\cal{A}$ under those inputs.  Without
loss of generality, we can order the intervals tested by $\cal{A}$ in the
sequence that $\cal{A}$ invokes the \tester on them in that execution. Let the
sequence of intervals be numbered from $1,\hdots, i$, for some value $i$.

Now, let us bound the probability of the event $E_i$, which is when
$\cal{A}$ tests intervals $1,\hdots ,i$ and fails. To do so, we
consider events associated with each invocation of \tester $j \in
[1,i]$.
Let $R_j$ be the event that $\cal{A}$ returns immediately after
invoking the \tester on the $j$-th interval. Let $\correctans{j}$ denote the
event that \tester returns a correct answer for the $j$-th interval.
If $E_i$ is true, then $\cal{A}$ terminates immediately after testing
the $i$-th interval and fails. This event happens only if two
conditions are met: First, $\cal{A}$ did not return immediately after
testing intervals $1,\hdots,i-1$; and second, $\cal{A}$ returns a
wrong answer at $i$-th interval and does terminate.
Therefore, we can conclude that the event $E_i = \wrongans{i} \cap
R_{i} \cap \noreturn{i-1} \cap \ldots \cap \noreturn{1}$. The
probability of failure $Pr[E_i]$ for each event $E_i$ is upper bounded
by $Pr[\wrongans{i}]$. 

Lastly, let $E$ be the total failure probability of $\cal{A}$.  We can
now use a union bound over possible failure events $E_1, \hdots, E_n$,
where $n$ is the maximum number of intervals $\cal{A}$ can possibly
test under any given input. Specifically:

\begin{align*}
Pr[E] = Pr[E_1 \cup E_2 \ldots \cup E_{n}] \leq \sum_{i=1}^{n}Pr[E_i]
\leq \sum_{i=1}^{n}Pr[\wrongans{i}]
\end{align*}

By analyzing $\sum_{i=1}^{n}Pr[\wrongans{i}]$ in the context of our
specific algorithm \bincert, we show that the quantity is bounded by $\delta$
(Lemma~\ref{lemma:adapt-sound}).
\end{proof}

\begin{lemma}
\label{lemma:adapt-sound}
Under any given input ($\theta, \error, \delta$), let $\correctans{i}$
be the event that $i$-th call made by \bincert to the \tester is
correct and let $n$ be the total number of calls to \tester by
\bincert. Then, $\sum_{i=1}^{n}Pr[\wrongans{i}] \leq \delta$.
\end{lemma}

\begin{proof}
We can upper bound the
number of proving intervals on the left by $n_l \leq 1 +
log\frac{\thresh}{\error}$. Similarly, for the right side of $\thresh
+ \error$, number of refuting intervals is $n_r \leq 1 +
log\frac{1-\thresh-\error}{\error}$. Lastly, there is only $1$ call to
the \tester on the interval $\error$ (line $19$,
Alg.~\ref{alg:adapt}). The total number of intervals tested in any one
execution of \bincert is at most $n = 3 + log\frac{\thresh}{\error} +
log\frac{1-\thresh-\error}{\error}$.
Each call to the \tester is done with confidence parameter $\delta_{min} =
\frac{\delta}{n}$, therefore by Lemma~\ref{lemma:tester}, the failure
probability of any call is at most $\frac{\delta}{n}$.  It follows that the
$Pr[\wrongans{i}] \leq n \cdot \frac{\delta}{n} =
  \delta$

\end{proof}

\section{Evaluation}
\label{sec:eval}
\begin{itemize}
\item {\em Scalability:} For a given timeout, what is the largest model that
	\tool and existing qualitative
	analysis tools can produce conclusive results.
\item {\em Utility in attack evaluations:} How does adversarial
  hardness computed with \codename compare with the efficacy of
  state-of-the-art attacks?
\item {\em Applicability to randomized models:} Can \tool certify
  properties of randomized DNNs?
\item {\em Performance.}  How many samples are needed by our new
  algorithms vis-a-vis the estimation approach (Section~\ref{sec:baseline})?
\end{itemize}

We implement our algorithms in a prototype tool called \tool and evaluate the
robustness of $38$ deterministic neural networks trained on $2$ datasets:
MNIST~\cite{lecun2010mnist} and ImageNet~\cite{ILSVRC15}.  For MNIST, we
evaluate on 100 images from the model's respective test set.  In the case of
ImageNet, we pick from the validation set as we require the correct label. 
Table~\ref{tab:benchmark_models} provides the size statistics of these models.
In addition, we evaluate the randomized model publicly provided by
\pixeldp~\cite{lecuyer2018certified}, which has a qualitative certificate of
robustness.

\begin{table}[]
\centering
\caption{Neural network architectures used in our evaluation.}
\label{tab:benchmark_models}
\resizebox{0.78\columnwidth}{!}{%
\begin{tabular}{|c|l|l|l|}
\hline
\textbf{Dataset} & \multicolumn{1}{c|}{\textbf{Arch}} & \multicolumn{1}{c|}{\textbf{Description}} & \multicolumn{1}{c|}{\textbf{\begin{tabular}[c]{@{}c@{}}\#Hidden\\ Units\end{tabular}}} \\ \hline
\multirow{6}{*}{MNIST(\bm)} & FFNN & 6-layer feed-forward & 3010 \\ \cline{2-4} 
 & convSmall & 2-layer convolutional & 3604 \\ \cline{2-4} 
 & convMed & 3-layer convolutional & 4804 \\ \cline{2-4} 
 & convBig & 6-layer convolutional & 34688 \\ \cline{2-4} 
 & convSuper & 6-layer convolutional & 88500 \\ \cline{2-4} 
 & skip & residual & 71650 \\ \hline
 \multicolumn{1}{|l|}{\multirow{5}{*}{ImageNet (\bmbig)}} & VGG16 & 16-layer convolutional & 15,086,080\\ \cline{2-4} 
\multicolumn{1}{|l|}{} & VGG19 & 19-layer convolutional & 16,390,656  \\ \cline{2-4} 
\multicolumn{1}{|l|}{} & ResNet50 & 50-layer residual & 36,968,684 \\ \cline{2-4} 
\multicolumn{1}{|l|}{} & Inception\_v3 & 42-layer convolutional & 32,285,184 \\ \cline{2-4} 
\multicolumn{1}{|l|}{} & DenseNet121 & 121-layer convolutional & 49,775,084 \\ \hline
\end{tabular}%
}
\end{table}

\paragraph{\eran Benchmark (\bm)}
Our first benchmark consists of $33$ moderate size neural networks
trained on the MNIST dataset.  These are selected to aid a comparison
with a state-of-the-art qualitative verification framework called
\eran~\cite{singh2018fast}. We selected all the models which \eran reported on.
These models range from 2-layer neural networks to 6-layer neural
networks with up to about $90K$ hidden units. We use the images used
to evaluate the \eran tool.

\paragraph{Larger Models (\bmbig)}
The second benchmark consists of $5$ larger deep-learning models pretrained on
ImageNet: VGG16, VGG19, ResNet50, InceptionV3 and DenseNet121. These 
models were obtained via the Keras framework with Tensorflow backend. These have
about $15-50M$ hidden units.

All experiments were run on GPU (Tesla V100-SXM2-16GB) with a timeout of $600$
seconds per image for the MNIST, $2000$ seconds for ImageNet models, and $3600$
seconds for the randomized \pixeldp model.

\subsection{Utility in Attack Evaluation}

\codename can be used to directly certify if the security analyst has
a threshold they want to check, for example, obtained from an external
specification. Another way to understand its utility is by relating
the quantitative bounds obtained from \codename with those reported by
specific attacks. When comparing the relative robustness of DNNs to
adversarial attacks, a common evaluation methodology today is to find
the minimum adversarial perturbation with which state-of-the-art
attacks can produce at least one successful adversarial example. If
the best known attacks perform worse on one DNN versus another, on a
sufficiently many test inputs, then the that DNN is considered
more robust.

\begin{table}[ht!]
  \centering
  \caption{Attack correlation for the PGD and C\&W attack for the models in BM1
and BM2 using Pearson's coefficient ($\rho$). For all, statistical significance
$p < 0.01$.}
\label{tab:attack-correlation}
\resizebox{0.8\columnwidth}{!}{%
\begin{tabular}{|l|c|c|}
\hline
\multicolumn{1}{|c|}{\textbf{Models}} & \textbf{$\rho$ (PGD)} & \textbf{$\rho$
(C\&W)} \\ \hline
convBigRELU\_\_DiffAI & 0.9617 & 0.7509 \\ \hline
convMedGRELU\_\_PGDK\_w\_0.1 & 0.8143 & 0.6686 \\ \hline
convMedGRELU\_\_PGDK\_w\_0.3 & 0.7699 & 0.6715 \\ \hline
convMedGRELU\_\_Point & 0.8461 & 0.982 \\ \hline
convMedGSIGMOID\_\_PGDK\_w\_0.1 & 0.8533 & 0.8903  \\ \hline
convMedGSIGMOID\_\_PGDK\_w\_0.3 & 0.9394 & 0.913 \\ \hline
convMedGSIGMOID\_\_Point & 0.9424 & 0.9605 \\ \hline
convMedGTANH\_\_PGDK\_w\_0.1 & 0.9521 & 0.9334 \\ \hline
convMedGTANH\_\_PGDK\_w\_0.3 & 0.9567 & 0.8718 \\ \hline
convMedGTANH\_\_Point & 0.7592 & 0.9817 \\ \hline
convSmallRELU\_\_DiffAI & 0.9504 & 0.5127 \\ \hline
convSmallRELU\_\_PGDK & 0.7803 & 0.6411 \\ \hline
convSmallRELU\_\_Point & 0.893 & 0.9816 \\ \hline
convSuperRELU\_\_DiffAI & 0.687 & 0.3856 \\ \hline
DenseNet-res & 0.7168  & 0.4879 \\ \hline
ffnnRELU\_\_PGDK\_w\_0.1\_6\_500 & 0.8932 & 0.9577 \\ \hline
ffnnRELU\_\_PGDK\_w\_0.3\_6\_500 & 0.7039 & 0.6496 \\ \hline
ffnnRELU\_\_Point\_6\_500 & 0.954 & 0.9788 \\ \hline
ffnnSIGMOID\_\_PGDK\_w\_0.1\_6\_500 & 0.8706 & 0.8955 \\ \hline
ffnnSIGMOID\_\_PGDK\_w\_0.3\_6\_500 & 0.9402 & 0.9201 \\ \hline
ffnnSIGMOID\_\_Point\_6\_500 & 0.8906 & 0.9489 \\ \hline
ffnnTANH\_\_PGDK\_w\_0.1\_6\_500 & 0.8156 & 0.9508 \\ \hline
ffnnTANH\_\_PGDK\_w\_0.3\_6\_500 & 0.9104 & 0.9485 \\ \hline
ffnnTANH\_\_Point\_6\_500 & 0.8998 & 0.8435 \\ \hline
mnist\_conv\_maxpool & 0.9664 & 0.9699 \\ \hline
mnist\_relu\_3\_100 & 0.9668 & 0.9448 \\ \hline
mnist\_relu\_3\_50 & 0.9702 & 0.9298 \\ \hline
mnist\_relu\_4\_1024 & 0.8945 & 0.9723 \\ \hline
mnist\_relu\_5\_100 & 0.7472 & 0.9629 \\ \hline
mnist\_relu\_6\_100 & 0.9845 & 0.9868 \\ \hline
mnist\_relu\_6\_200 & 0.9699 & 0.9412 \\ \hline
mnist\_relu\_9\_100 & 0.979 & 0.9725 \\ \hline
mnist\_relu\_9\_200 & 0.8165 & 0.9652 \\ \hline
ResNet50 & 0.7929 & 0.6932 \\ \hline
skip\_\_DiffAI & 0.7344 & 0.6298 \\ \hline
VGG16 & 0.8064 & 0.8297 \\ \hline
VGG19 & 0.7224 & 0.7335 \\ \hline
Inception-v3 &  0.5806 & 0.4866 \\ \hline
\end{tabular}%
}
\end{table}

\codename offers a new capability: we can measure the ratio of
adversarial samples within a specified perturbation bound of a given
test input $\vec{x}$ (see
Section~\ref{sec:application}). Specifically, we can compute the {\em
  adversarial density} by uniformly sampling in the $L_p$ ball of
$\vec{x}$, and checking if the ratio of adversarial samples is very
small (below $\theta=10^{-3}$). By repeating this process for
different perturbations bounds, we empirically determine the {\em
  adversarial hardness} (or \advhard)---the largest perturbation bound
below which \codename certifies the adversarial density to be that
small (returns \yes) and above which \codename does not (returns \no).
We use density threshold $\theta=10^{-3}$, error tolerance
$\eta=10^{-3}$, and confidence parameter $\delta=0.01$.

\codename computes the adversarial hardness with black-box access.  As
a comparison point, we evaluate against two {\em white-box}
attacks --- \pgd~\cite{madry2017towards} for $L_{\infty}$ and
\carlini~\cite{carlini2017towards} for
$L_2$ implemented in CleverHans~\cite{papernot2018cleverhans} (v3.0.1) --- 
$2$ prominent attacks that are recommended for the $L_p$ norms we
consider~\cite{carlini2019evaluating}.  White-box attacks enable the
attacker complete access to internals; therefore, they are 
more powerful than black-box attacks today.  Both \pgd and \carlini are
gradient-based adversarial attacks. For \pgd, we perform $30$ attacks on
different values of $\epsilon$ to identify the minimum value that an adversarial
input can be identified. For \carlini, we identify the best $\epsilon$ it
is able to identify for a given amount of resource (iterations).

Our main empirical result in this experiment is that \advhard is {\em strongly correlated} with
$\epsilon_{min}$. Figure~\ref{fig:pgd-correlation} and
Figure~\ref{fig:carlini-correlation} show the correlation visually for
two models: it shows that the perturbation bounds found by these two
separate attacks are different, but both correlate with the
adversarial hardness of the certification instance produced by
\codename.
The Pearson correlation for all models is reported in
Table~\ref{tab:attack-correlation} for all cases where the compared
white-box attacks are successful
The average Pearson correlation between the perturbation found by
\pgd, \pgdeps, and \advhard over all models is $0.858$ and between
the perturbation found by \carlini, \carlinieps, and \advhard is
$0.8438$.  We take $25$ images per model to calculate the
correlation. The significance level is high (p-value is below $0.01$
for all cases).

Recall that \codename is sound, so the estimate of adversarial
hardness \advhard is close to the ground truth with high probability
($99\%$).  This metric is an attack-agnostic metric, computed by
uniform sampling and {\em without} white-box access to the model. The
strong correlation shows that \pgd and \carlini
attacks find smaller $\epsilon_{min}$ for easier certification
instances, and larger $\epsilon_{min}$ for harder instances. This
suggests that when comparing the robustness of two models, one can
consider adversarial hardness as a useful {\em attack-agnostic} metric,
complementing evaluation on specific attacks.

\begin{figure*}
  \centering
  \begin{minipage}[b]{0.31\textwidth}
      \centering
  \resizebox{.99\textwidth}{!}{%
\begin{tikzpicture}
\begin{axis}[
  const plot,
  stack plots=y,
  area style,
  enlarge x limits=false,
  legend style={at={(0.5,-0.20)},
    anchor=north,legend columns=-1},
  ylabel={$\epsilon$ value},
  xlabel={Image Id},
  ]
  \addplot coordinates
    {(1, 0.09243)
     (2, 0.06162)
     (3, 0.09243)
     (4, 0.12324)
     (5, 0.06162)
     (6, 0.09243)
     (7, 0.06162)
     (8, 0.03081)
     (9, 0.03081)
     (10, 0.06162)
     (11, 0.06162)
     (12, 0.12324)
     (13, 0.09243)
     (14, 0.09243)
     (15, 0.09243)
     (16, 0.06162)
     (17, 0.06162)
     (18, 0.09243)
     (19, 0.006162)
     (20, 0.06162)
     (21, 0.06162)
     (22, 0.06162)
     (23, 0.09243)
     (24, 0.09243)
     (25, 0.06162)
     } 
    \closedcycle;
  \addplot coordinates
    {(1, 0.6797)
     (2, 0.4531)
     (3, 0.5869)
     (4, 0.7206)
     (5, 0.4531)
     (6, 0.5869)
     (7, 0.3293)
     (8, 0.3193)
     (9, 0.2575)
     (10, 0.4531)
     (11, 0.4531)
     (12, 0.6587)
     (13, 0.5869)
     (14, 0.6797)
     (15, 0.5869)
     (16, 0.3293)
     (17, 0.4531)
     (18, 0.5869)
     (19, 0.101)
     (20, 0.4531)
     (21, 0.4531)
     (22, 0.4531)
     (23, 0.5869)
     (24, 0.6797)
     (25, 0.3293)
    }
    \closedcycle;
  \legend{\pgdeps, \advhard}
\end{axis}
\end{tikzpicture}
}
\captionof{figure}{Correlation graph between $L_{\infty}$ bounds provided by \codename and \pgd for a
fully connected feedforward with \texttt{sigmoid} (FFNN) on MNIST.}
\label{fig:pgd-correlation}
  \end{minipage}%
  \hfill
  \begin{minipage}[b]{0.31\textwidth}
      \centering
  \resizebox{.99\textwidth}{!}{%
\begin{tikzpicture}
\begin{axis}[
  const plot,
  stack plots=y,
  area style,
  enlarge x limits=false,
  legend style={at={(0.5,-0.20)},
    anchor=north,legend columns=-1},
  ylabel={$\epsilon$ value},
  xlabel={Image Id},
  ]
  \addplot coordinates
    {(1, 1.484827)
     (2, 1.147088)
     (3, 1.370236)
     (4, 1.698023)
     (5, 0.933585)
     (6, 1.411428)
     (7, 0.762637)
     (8, 0.735179)
     (9, 0.630139)
     (10, 1.07338)
     (11, 1.290346)
     (12, 1.710903)
     (13, 1.33126)
     (14, 1.803899)
     (15, 1.489307)
     (16, 0.67626)
     (17, 1.240902)
     (18, 1.409377)
     (19, 0.114788)
     (20, 1.080263)
     (21, 0.994062)
     (22, 1.153031)
     (23, 1.288433)
     (24, 1.494648)
     (25, 0.692685)
     } 
    \closedcycle;
  \addplot coordinates
    {(1, 9.6328)
     (2, 7.2188)
     (3, 8.1563)
     (4, 11.1875)
     (5, 6)
     (6, 8.9063)
     (7, 4.625)
     (8, 4.4063)
     (9, 4.0313)
     (10, 6.5625)
     (11, 6.75)
     (12, 9.8125)
     (13, 7.5)
     (14, 9.9843)
     (15, 8.3438)
     (16, 4.625)
     (17, 6.8438)
     (18, 8.625)
     (19, 0.75)
     (20, 7.3125)
     (21, 6.1875)
     (22, 6.4688)
     (23, 9.0703)
     (24, 10.4062)
     (25, 4.625)
    }
    \closedcycle;
  \legend{\carlinieps, \advhard}
\end{axis}
\end{tikzpicture}
}
\captionof{figure}{Correlation graph between $L_{2}$ bounds provided by \codename and
\carlini for a fully connected feedforward (FFNN) with \texttt{sigmoid} on
MNIST.}
\label{fig:carlini-correlation}
  \end{minipage}
  \hfill
  \begin{minipage}[b]{0.33\textwidth}
      \centering
  \resizebox{.99\textwidth}{!}{%
\begin{tikzpicture}

\begin{axis}[ybar stacked,
  xlabel=Perturbation Size,
  ylabel=Verified Robustness (\%),
  ymax=1.05,
  ymin=0.0,
  legend style={at={(0.5,-0.20)},
  anchor=north,legend columns=-1}]
	\addplot coordinates
		{(0.01,0.943185346165787)
      (0.03,0.8078236572493015)
    (0.05,0.7506985408258305)
    (0.08,0.6563179136914001)
    (0.1,0.6395529338714685)
    (0.13, 0.5644209872710338)
    (0.15, 0.6159577770878609)
    (0.18, 0.4843216392424713)
    (0.2,0.37069233157404535)
    (0.23,0.45700093138776776)
    (0.25,0.5569698851288419)};
	\addplot coordinates
  {(0.01,0.056814653834212976)
    (0.03,0.1620614715926731)
    (0.05,0.22011797578391804)
    (0.08, 0.2694815274759392)
    (0.1, 0.29804408568767465)
    (0.13, 0.3548587395218876)
    (0.15,0.32629618131015214)
    (0.18,0.3772120459484632)
    (0.2, 0.47842285004656937)
    (0.23, 0.40732691710648866)
  (0.25,0.34616578702266376)};
	\addplot coordinates
  {(0.01,0.0)
    (0.03,0.03011487115802546)
    (0.05, 0.029183483390251473)
    (0.08,0.07420055883266066)
    (0.1,0.062402980440856876) 
    (0.13, 0.08072027320707854)
    (0.15,0.05774604160198696)
    (0.18,0.13846631480906552)
    (0.2,0.15088481837938528)
    (0.23,0.13567215150574355)
    (0.25,0.09686432784849426)};
\legend{\codename \yes, \codename \no, \codename \none}
	\end{axis}
  \begin{axis}[axis y line=none,
    ymax=1.05,
    ymin=0.0]
\addplot[thick,color=black,mark=x] coordinates {
	(0.01,0.9046879850977957)
	(0.03, 0.5811859670909656)
	(0.05,0.43682086308599816)
	(0.08,0.26911618669314796)
	(0.1,0.26911618669314796)
  (0.13,0.06874557051736357)
	(0.15,0.07743216412971542)
	(0.18,0.05890138980807412)
  (0.2,0.04401058901389808)
  (0.23,0.026295731707317072)
  (0.25,0.019523494374586368)
};
\legend{\eran}
\end{axis}
\end{tikzpicture}
}
\captionof{figure}{\codename and \eran verified robustness per perturbation size for \bm
for threshold $\thresh=0.0001$, error $\error=0.001$ and confidence $\delta=0.01$.}
\label{fig:eran_provero}
  \end{minipage}

\end{figure*}

\subsection{Scalability}
We test \tool on $38$ models, which range from $3K - 50M$ hidden
units. We select $100$ input images for each model and retain all
those inputs for which the model correctly classifies.
We tested $11$ perturbation size ($L_{\infty}$) from $0.01$ to $0.25$ for \bm
and $4$ perturbation size ($L_2$) from ${2/255}$ to $14/255$ for \bmbig.
This results in a total of $36971$ test images for $38$ models.  We run each
test image with \codename with the following parameters:
$(\thresh=0.0001,\error=0.001$ and $\thresh=0.01,\error=0.01,\delta=0.01)$ (for \bm)
$(\thresh=0.001,\eta=0.001$ and $\thresh=0.01,\eta=0.01,\delta=0.01)$ (for \bmbig).
We find that \tool scales producing answers within the timeout of $600$ seconds
for \bm and $2000$ seconds for \bmbig per test image.
 Less than $2\%$ input cases for \bmbig and less than $5\%$ for \bm return
 'None', i.e., \codename cannot certify conclusively that there are less or more
 than than the queried thresholds. For all other cases, \codename provides a
 \yes or \no results.

 As a comparison point, we tried to compare with prior work on quantitative
 verification~\cite{BSSMS19}, prototyped in a tool called NPAQ. While NPAQ
 provides direct estimates rather  than the \yes or \no answers \codename
 produces, we found that NPAQ supports only a sub-class of neural networks
 (BNNs) and of much smaller size. Hence, it cannot
 support or scale for any of the models in our benchmarks, \bm and \bmbig.

 Secondly, we tested \eran which is considered as the most
scalable qualitative verification tool. \eran initially was not able to parse these
models. After direct correspondence with the authors of \eran, the
authors added support for requisite input model formats. After
applying these changes, we confirmed that the current implementation
of \eran time out on all the \bmbig models. \codename finishes
on these within a timeout of $2000$ seconds.

We successfully run \eran on \bm, which are smaller benchmarks that
\eran reported on.
There are total of $4$ analyses in \eran. We evaluate on the \deepzono
and \deeppoly domains but for the \deeppoly, on our evaluation
platform, \eran runs out of memory and could not analyze the neural
networks in \bm.  The remaining analyses, \refinezono and \refinepoly
are known to achieve or improve the precision of the \deepzono or
\deeppoly domain at the cost of larger running time by calling a
mixed-integer programming solver~\cite{singh2018boosting}.  Hence, we
compare with the most scalable of these $4$ analyses, namely the \deepzono
domain.

Figure~\ref{fig:eran_provero} plots the precision of \eran against
\codename for all $11$ perturbation sizes.  We find that for a
perturbation size of more than $0.05$, \eran's results are
inconclusive, i.e., the analysis reports neither \yes nor \no for more
than $50\%$ of the inputs, likely due to imprecision in
over-approximations of the analysis.
Figure~\ref{fig:eran_provero} shows that the verified models reduces
for higher perturbations $\epsilon$. This is consistent with the
findings in the \eran paper: \eran either takes longer or is more
imprecise for non-robust models and higher $\epsilon$
values~\cite{singh2018boosting}.
In all cases where \eran is inconclusive, \tool successfully finishes
within the $600$ second timeout for all $35431$ test images and values of
$\epsilon$. In above $95\%$ of these cases, \codename produces
high-confidence \yes or \no results.

As a sanity check, on cases where \eran conclusively outputs a \yes, \tool also
reports \yes.  With comparable running time, \codename is able to obtain
quantitative bounds for all perturbation sizes.
From these experimental results, we conclude \codename is a complementary
analysis tool that can be used to quantitatively certify larger models and for
larger perturbation sizes, for which our evaluated qualitative verification
framework (\eran) is inconclusive.  To the best of our knowledge, this is the
first work to give any kind of sound quantitative guarantees for such large
models.

\subsection{Applicability to Randomized DNNs}
\label{sec:pixeldp-eval}

So far in our evaluation we have focused on deterministic DNNs,
however, \tool can certify the robustness of randomized DNNs.  To
demonstrate this generality, we take a model obtained by a training
procedure called \pixeldp that adds differentially private noise to
make the neural network more robust.  The inference phase of a
\pixeldp network uses randomization: instead of picking the label with
the maximum probability, it samples from the noise layer and
calculates an expectation of the output value. \pixeldp also produces
a certified perturbation bound for which it guarantees the model to be
robust for a given input image.
Note that qualitative verification tools such as \eran require
white-box access and work with deterministic models, so they would not
be able to verify the robustness of randomized \pixeldp DNNs.

We contacted the authors to obtain the models used in
\pixeldp~\cite{lecuyer2018certified}.  The authors pointed us to the
\pixeldp ResNet28-10 model trained on
CIFAR10
as the main representative of the technique.
We randomly select $25$ images in the CIFAR10 dataset and for each
image we obtain the certified perturbation bound $\epsilon_{pixeldp}$
produced by \pixeldp itself. We configure \pixeldp to internally
estimate $\epsilon_{pixeldp}$ using $25$ samples from the noise layer
as recommended in their paper. This bound, $\epsilon_{\pixeldp}$, is
the maximum bound for which \pixeldp claims there are no adversarial
samples within the $L_2$ ball.

We use \tool to check the certificate $\epsilon_{\pixeldp}$ produced
by \pixeldp, using the following parameters
$\thresh=0.01,\error=0.01,\delta=0.01$. \tool reports \yes, implying
that the model has adversarial density under these bounds.  Under the
same threshold $\thresh=10^{-2},\eta=10^{-2}$ we tested for larger
perturbation sizes: from $\epsilon=0.1$ to $0.25$. 

Our findings in this experiment are that \codename can certify low
adversarial density for perturbation bounds much larger than the
qualitative certificates produced by \pixeldp.
In particular, \codename certifies that for $\epsilon=5
\times\epsilon_{\pixeldp}$, the \pixeldp model has less than
$10^{-2}$ adversarial examples with confidence at least $99\%$. 
\codename offers complimentary quantitative estimates of robustness for
\pixeldp.

\subsection{Performance}

Our estimation solution outlined in Section~\ref{sec:baseline} applies
Chernoff bounds directly. For a given precision parameter $\error$, it
requires a large number of samples, within a factor of $1/\eta^2$.
While we do not escape from this worst case, we show that our
proposed algorithms improve over the estimation baseline empirically.
To this end, we record the number of samples taken for each test image
and compare it to the number of samples as computed for the estimation
approach.

We find that \tool requires  $27\times$ less samples
than the estimation approach for values of $\error=0.01,\thresh=0.01$ and
$687\times$ less samples for values of $\error=0.001,\thresh=0.0001$ for \bm.
For larger models in \bmbig and values of $\error=0.001,\thresh=0.001$, \tool
requires $74\times$ less samples than the estimation solution and
$\error=0.01,\thresh=0.01$.

In our implementation, we use a batch-mode to do the inference for
models in \bm and \bmbig, speeding up the running time of the sampling
process by a factor of $3\times$. This leads to average times of
$29.78$ seconds per image for \bm for a batch size of $128$. For the
models in \bmbig we used different batch sizes so we report the
average time per sample as $0.003$ seconds which can be used to derive the
running time based.
For ($\thresh=0.0001,\eta=0.001$) and all values of
$\epsilon$, the average number of samples over all images is $80,424$ (median value is
$83,121$, standard deviation is $14,303$).  For VGG16 the average
number of samples is around $657,885$, for VGG19 it is around
$644,737$ and for InceptionV3 $664,397$ samples, respectively, for
$\thresh=0.001,\error=0.001$. For both $\thresh=0.01,\error=0.01$ and
$\thresh=0.001,\error=0.001$, for VGG16 the average number of samples
is $737,297$, for VGG19 it is $722,979$ and for InceptionV3 $744,792$
average number of samples, resulting in about $74\times$ less samples
than the estimation approach.

For the \pixeldp model, \codename takes $20,753$ samples. Since
\pixeldp models internally take $25$ samples from the noise layer, the
time taken for inference on one given input itself is higher.
\codename reports an average running time per test image of roughly
$1,500$ seconds.

\section{Related Work}

Qualitative verification methodologies have sought to exploit the advances in
combinatorial solving, thereby consisting of satisfiability modulo theories
(SMT) solvers-based
approaches~\cite{huang2017safety,katz2017reluplex,katz2019marabou,ehlers2017formal,narodytska2017verifying}
or integer linear programming
approaches~\cite{tjeng2017evaluating,lin2019robustness,cheng2017maximum}.  Despite singficiant progress over the years,
 the combinatorial solving-based approaches do not
scale to deep neural networks.

Consequently, techniques to address scalability often sacrifice completeness:
abstract interpretation-based
techniques~\cite{gehr2018ai2,singh2018fast,singh2018boosting,singh2019abstract}
are among the most scalable verification procedures that over-approximate the
activation functions with abstract domains. Similarly, optimization-based
approaches~\cite{wong2018provable,raghunathan2018certified,dvijotham2018dual}
produce a certificate of robustness for a given input: a lower bound on the
minimum perturbation bound $\epsilon_{min}$ such that there are no adversarial
examples within that a $\epsilon_{min}$ ball. These techniques are, however,
incomplete and catered to only ReLU activations~\cite{wong2018provable} and
fully-connected layers~\cite{zhang2018efficient,gopinath2019property}. 

A promising line of incomplete techniques has been proposed employing two
complementary techniques: Lipschitz computation and linear approximations. Hein
and Andriushchenko~\cite{hein2017formal} propose an analytical analysis based on
the Lipschitz constant, but the approach assumes that a differentiable
activation function (thus excluding ReLU activations) and can handle only two
layers.   Boopathy et al~\cite{boopathy2019cnn}, Weng et
al.~\cite{weng2018towards}, Zhang et al~\cite{zhang2018efficient} have further
improved the scalability of these techniques but they are
still limited to  $80,000$ hidden units. While the
lower and upper bounds are sound, their tightness is not guaranteed.  In another
line of work, Weng et al.~\cite{weng2018evaluating} employ extreme value theory
to estimate a lower bound on $\epsilon_{min}$ albeit without theoretical
guarantees of the soundness of the obtained bounds.
\codename differs fundamentally from classical verification approaches in our
focus on the development of a quantitative verification framework with rigorous
guarantees on computed estimates.

Baluta et al.~\cite{BSSMS19} propose a quantitative framework that outputs
estimates of how often a property holds.  
network into a logical formula and use model counting tools. 
Their approach, however,  is instantiated to a specific type of neural networks, namely binarized
neural networks, and its scalability is limited to networks with around $50,000$
parameters.  In contrast,  \codename scales to DNNs with millions of parameters
while preserving the formal guarantees of the white-box approach.
Webb et al.~\cite{webb2018statistical} proposed a Monte Carlo-based approach for
rare events to estimate the ratio of adversarial examples.   In this work, we
take a {\em property testing} approach wherein we ask if the proportion of
inputs that violate a property is less than a given threshold (in which case we
say yes) or it is $\eta$-far from the threshold (in which case we say no). While
both approaches are black-box and rely on sampling, \codename returns a yes or
no answer with user-specified high confidence, whereas the statistical approach
proposed by Webb et al. does not provide such guarantees. A related area is
statistical model
checking~\cite{larsen2014statistical,kwiatkowska2007stochastic,kwiatkowska2009prism}
that relies on sampling schemes~\cite{wald1945sequential} to accept a given
hypothesis. \codename algorithm is similar to a sequential sampling
plan~\cite{younes2005verification} but \codename's insight is that each test
checks ``cheaper'' hypotheses in a binary-search manner. In particular, in case of {\codename}, 
we design hypothesis in a manner that more {\em expensive} hypothesis tests are delayed as much as possible.

\section{Conclusion}

\codename introduces an algorithm for verifying quantitative
properties of neural networks, assuming only black-box access, and
with much better test sample complexity than compared baselines. Our
algorithm offers, in particular, a powerful new attack-agnostic way of
evaluating adversarial robustness for deep neural networks.

\section*{Acknowledgment}

This work was supported by Crystal Centre at National University of Singapore,
Singapore, National Research Foundation Singapore under its NRF Fellowship
Programme [NRF-NRFFAI1-2019-0004 ] and AI Singapore Programme [AISG-RP-2018-005],
and NUS ODPRT Grant [R-252-000-685-13]. Prateek Saxena is supported by a research award
from Google. Part of the computational work for this article was performed on
resources of the National Supercomputing Centre, Singapore and National
Cybersecurity R\&D Lab, Singapore.

\bibliographystyle{IEEEtranS}
\bibliography{IEEEabrv,paper}

\end{document}